\documentclass[conference]{IEEEtran}
\IEEEoverridecommandlockouts
\usepackage{cite}
\usepackage{amsmath,amssymb,amsfonts}
\usepackage{algorithmic}
\usepackage{graphicx}
\usepackage{textcomp}
\usepackage{xcolor}
\usepackage{layout}
\usepackage{subfig}
\usepackage{hyperref}
\usepackage{multirow}
\usepackage{booktabs}
\usepackage{colortbl}
\usepackage{xspace}
\newcommand{\ie}{i.e.\@\xspace}
\definecolor{graybg}{rgb}{0.9, 0.9, 0.9}

\def\BibTeX{{\rm B\kern-.05em{\sc i\kern-.025em b}\kern-.08em
    T\kern-.1667em\lower.7ex\hbox{E}\kern-.125emX}}

\begin{document}

\title{Enhancing Event-based Object Detection with Monocular Normal Maps}
\author{
\IEEEauthorblockN{
Mingjie Liu\textsuperscript{1,*}, 
Hanqing Liu\textsuperscript{1,*}, 
Luoping Cui\textsuperscript{1}, and 
Chuang Zhu\textsuperscript{1,\dag}
}
\vspace{0.15cm}
\IEEEauthorblockA{\textsuperscript{1}\textit{School of Artificial Intelligence, Beijing University of Posts and Telecommunications, Beijing, China}}
\IEEEauthorblockA{\texttt{\{LMJ, hanqingliu, lpcui, czhu\}@bupt.edu.cn}}
\thanks{*Equal contribution.}
\thanks{\textsuperscript{\dag}Corresponding author.}
}

\maketitle

\begin{abstract}
Object detection in autonomous driving is frequently compromised by complex illumination. While event cameras offer a robust solution, they are susceptible to sudden contrast changes such as reflections which often trigger dense, misleading event signals.  
To overcome this, we leverage RGB-derived surface normal maps as explicit geometric constraints. Crucially, even when RGB degrades, they preserve low-frequency structural priors that effectively assist in event-based detection. 
Consequently, we present \textbf{NRE-Net}, a trimodal framework that integrates structural priors from surface \textbf{N}ormal maps, appearance context from \textbf{R}GB images, and high-frequency dynamics from \textbf{E}vents. The Adaptive Dual-stream Fusion Module (ADFM) first aligns geometric and appearance cues, followed by the Event-modality Aware Fusion Module (EAFM) which selectively integrates event dynamics.
Extensive evaluations on DSEC-Det-sub and PKU-DAVIS-SOD demonstrate that incorporating geometric priors yields an additional 3.0\% AP$_{\text{50}}$ gain over dual-modal baselines, while our approach consistently outperforms fusion methods such as SFNet (+2.7\%) and SODFormer (+7.1\%).

\end{abstract}


\section{Introduction}
\label{sec:intro}

\begin{figure}[htbp!]
    \centering
    \includegraphics[trim=5.8cm 25cm 5cm 1cm, clip, width=0.45\textwidth]{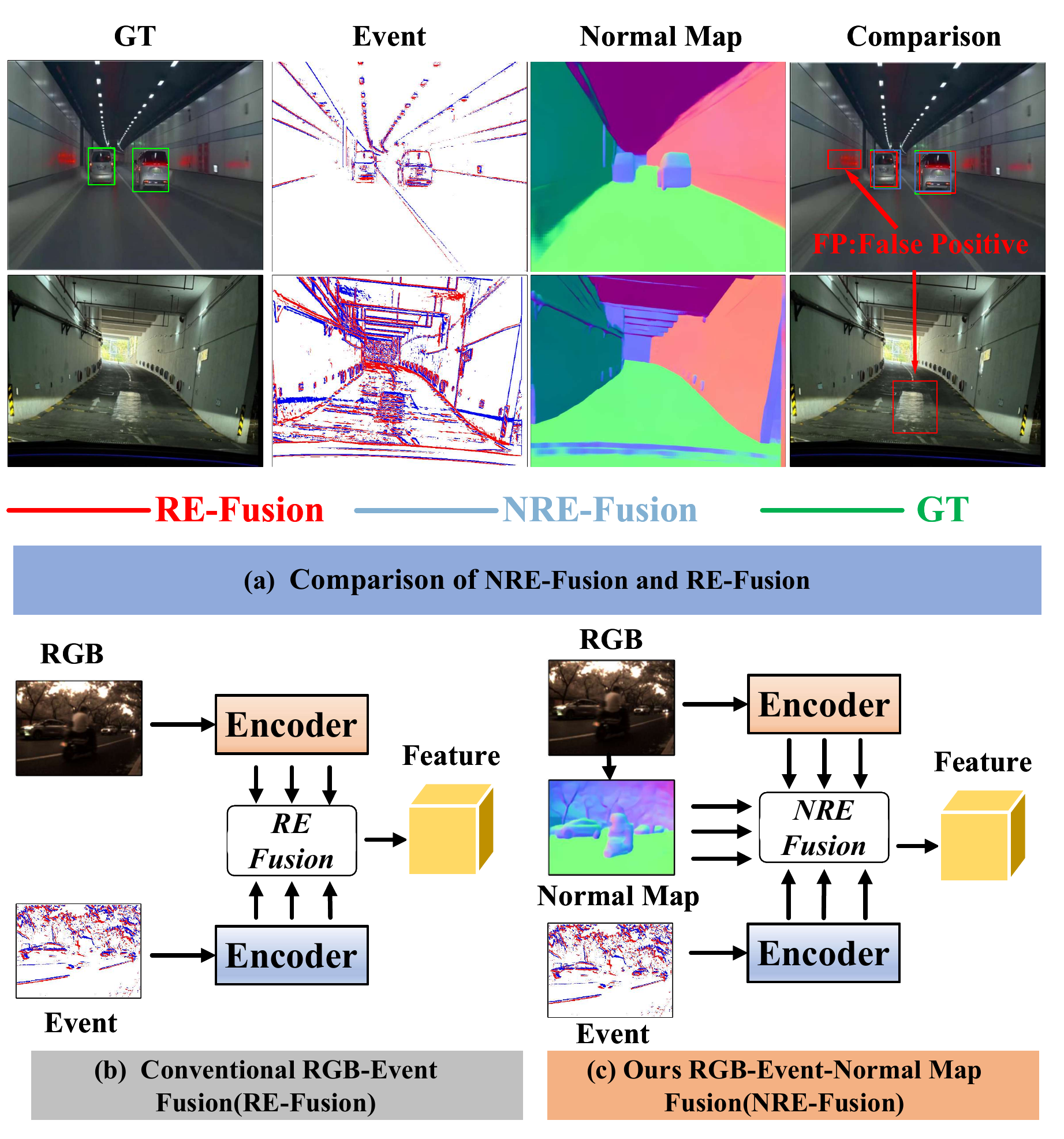}

    \caption{\textbf{Motivation. }While event cameras excel in adverse lighting, they are susceptible to dense, misleading event signals triggered by sudden contrast changes such as reflections. Our approach leverages surface normal maps as stable, low-frequency structural cues to enhance event-based detection performance.}
    \label{fig:introduction}

\end{figure}

Object detection remains one of the most fundamental yet challenging tasks in computer vision~\cite{zhao2024detrs}. Leveraging deep learning techniques, recent approaches have shown remarkable performance under controlled conditions using RGB images~\cite{cui2026peod}. However, when deployed in real-world autonomous driving scenarios, these methods often deteriorate sharply under complex illumination variations, as the limited dynamic range of standard RGB sensors fails to capture crucial visual cues~\cite{liu2024enhancing}.

Event cameras have gained prominence in such challenging environments. Unlike traditional frame-based sensors, event cameras output pixel-level brightness changes asynchronously, boasting a high dynamic range (over 120 dB) and low latency~\cite{liu2024enhancing}. While intrinsically robust to diverse lighting environments, event cameras introduce a new set of challenges: they are highly sensitive to sudden contrast changes such as reflections on tunnel walls or road surfaces. As illustrated in Fig.~\ref{fig:introduction} (a), these distracting reflections often generate strong event signals that mimic genuine obstacles, triggering false detections and potentially leading to dangerous ``Phantom Braking'' scenarios~\cite{berge2024phantom}.

To address this perceptual ambiguity, we introduce surface normal maps as an explicit geometric constraint. Unlike depth maps, which provide absolute distance but can be ambiguous regarding surface reflections (as reflections possess valid depth values), surface normals explicitly encode local surface orientation~\cite{bae2024rethinking}. In driving scenes, reflections typically align with the planar surface (e.g., the road), whereas physical obstacles exhibit vertical orientations perpendicular to the ground~\cite{So2022Analysis}.

We propose to estimate normal maps directly from monocular RGB images to assist event-based object detection. This design choice is driven by two main considerations. First, it facilitates auxiliary detection assistance without necessitating additional sensor hardware. Second, while reconstructing RGB images using event information might yield higher-quality normal maps, the resulting increase in model complexity yields only marginal improvements in detection performance\cite{liang2024towards}. 
Although it might be argued that RGB degradation in low-light conditions could impair normal estimation, our experiments suggest that low-frequency structural priors (e.g., the planarity of the road) are generally preserved. Crucially, our framework exploits these stable structural geometries rather than the high-frequency texture details that are often corrupted by adverse illumination. By integrating these complementary cues, we aim to resolve the ambiguities that plague purely appearance-based or motion-based methods.

\begin{figure*}[htbp]
    \centering
    \includegraphics[trim=2cm 5cm 1cm 2.7cm, clip, width=0.94\textwidth, height=8.5cm]{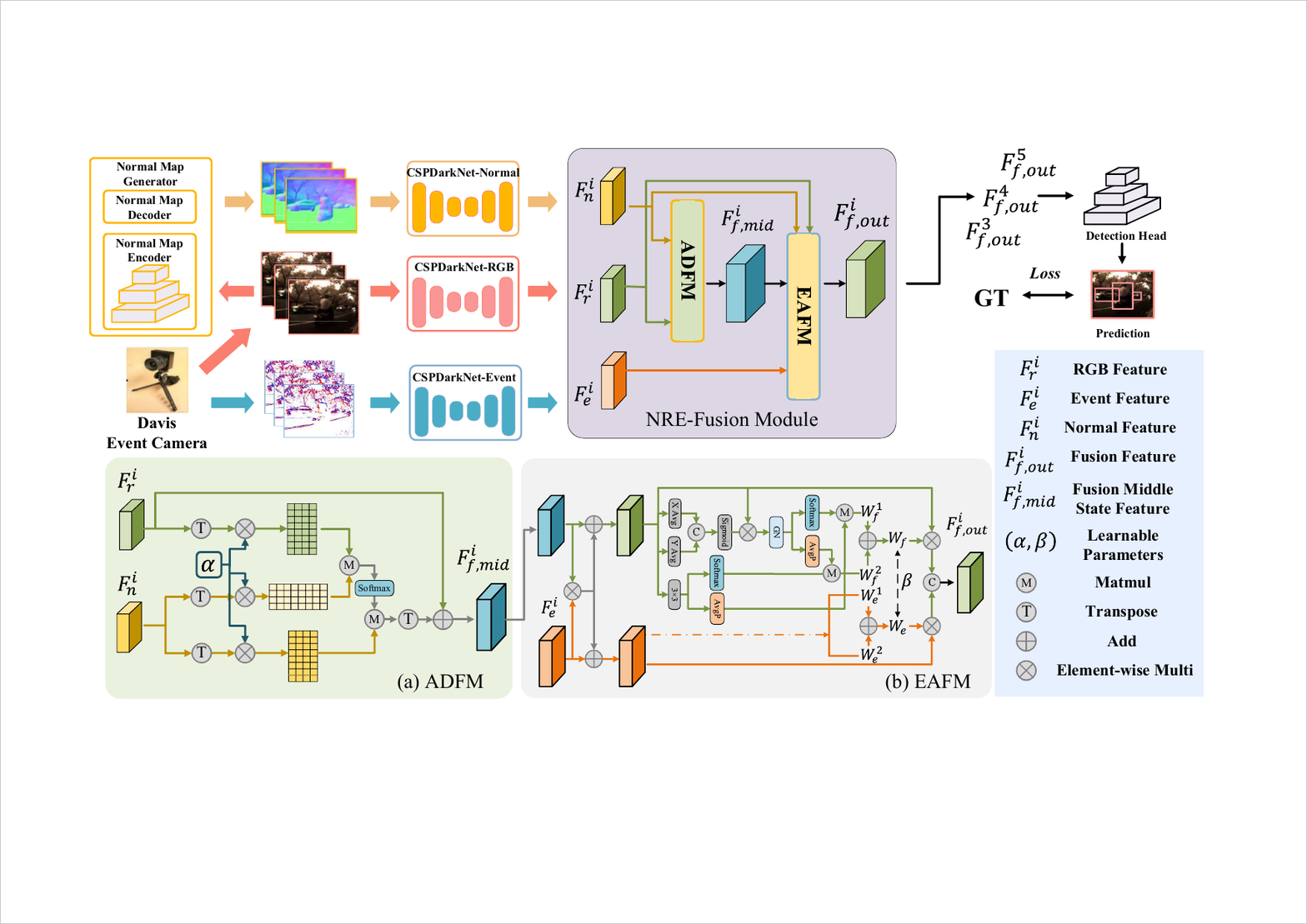}
    \caption{Overview of the proposed NRE-Net. (a) The Adaptive Dual-stream Fusion Module (ADFM) uses attention mechanisms to weigh the reliability of RGB and geometric features. (b) The Event-modality Aware Fusion Module (EAFM) integrates dynamic event signals. This hierarchical design ensures robust detection even under single-modality degradation.}
    \label{fig:framework}
\end{figure*}

In this paper, we present NRE-Net, a novel trimodal framework that synergistically fuses: (1) RGB images for appearance context, (2) Event streams for high dynamic range dynamics, and (3) Monocular Normal Maps for geometric consistency. Fig.~\ref{fig:framework} illustrates our architecture.
In summary, our key contributions are:
\begin{itemize} 
    \item We introduce monocularly predicted normal maps as structural geometries to assist in resolving perceptual ambiguities in event-based object detection.
    
    \item We design a robust hierarchical fusion architecture comprising two specialized modules: the Adaptive Dual-stream Fusion Module (ADFM) which synergizes geometric and appearance cues via cross-attention, and the Event-modality Aware Fusion Module (EAFM) which integrates high dynamic range event data while suppressing noise.

    \item We conduct extensive evaluations on the DSEC-Det-sub~\cite{Gehrig21ral} and PKU-DAVIS-SOD~\cite{li2023sodformer} datasets. Incorporating explicit geometric priors yields a 3.0\% AP$_{\text{50}}$ gain over standard dual-modal baselines, while our hierarchical fusion mechanism improves performance by 3.2\% compared to naive integration strategies.

\end{itemize}

\section{Related Work}

\subsection{Surface Normal Estimation}
Surface normals provide essential geometric cues for scene understanding \cite{fu2024geowizard}. While early methods relied on depth sensors or strict photometric assumptions \cite{Hashimoto2019Uncalibrated}, recent deep learning advancements have enabled robust estimation directly from monocular RGB images. State-of-the-art approaches, such as JointNet \cite{zhang2023jointnet} and Wonder3D \cite{long2024wonder3d}, leverage generative priors to ensure geometric consistency. Notably, DSINE \cite{bae2024rethinking} achieves superior generalization by explicitly modeling per-pixel ray directions and relative surface rotations. However, the integration of such geometric priors into object detection frameworks particularly for resilience against adverse lighting remains underexplored. We address this gap by incorporating monocular normals as a complementary geometric modality alongside RGB and event streams.

\subsection{Neuromorphic Object Detection}
Conventional RGB-based detectors often falter under extreme lighting or rapid motion. While neuromorphic event cameras offer a robust alternative due to their high dynamic range and low latency \cite{zhou2023rgb}, single-modality event methods inherently lack detailed texture and static semantic context.

To mitigate these limitations, multi-modal fusion has become a mainstream research direction. Early works, such as RENet \cite{zhou2023rgb} and EOLO \cite{cao2024chasing}, introduced calibration and symmetric fusion mechanisms to dynamically balance RGB and event modalities. SFNet \cite{liu2024enhancing} further incorporated structural awareness to enhance robustness against illumination variations. More recently, architectures leveraging streaming Transformers \cite{li2023sodformer} and asynchronous graph representations \cite{Li2025ACGR} have been proposed to exploit rich temporal cues for high-speed detection.
Distinct from these approaches, which primarily focus on appearance or temporal fusion, our framework introduces monocular surface normal maps as an explicit scene structural prior to assist event-based object detection. By integrating these, we achieve robustness in adverse lighting conditions where texture information is unreliable.

\subsection{Multi-Modal Fusion}
Multi-modal fusion is pivotal for enhancing perceptual robustness in adverse lighting. While \emph{late fusion} strategies \cite{li2019event} consolidate final predictions via probabilistic voting, they often fail to exploit deep cross-modal synergy. In contrast, \emph{feature-level fusion} integrates representations within a shared latent space. Recent approaches leverage attention mechanisms, multi-stage blending, or depth guidance \cite{zhou2021mffenet, palladin2024samfusion} to capture inter-modal interactions. However, these methods often overlook global spatial correlations or lack explicit mechanisms to handle severe modality-specific degradation.

To address these limitations, we propose a hierarchical fusion strategy comprising two specialized modules: the Adaptive Dual-stream Fusion Module (ADFM) and the Event-modality Aware Fusion Module (EAFM). ADFM utilizes global cross-attention to synergize RGB appearance with geometric priors from normal maps, while EAFM adaptively integrates high-dynamic-range event features using learned spatial weighting. Together, they effectively mitigate degradation and enhance detection stability under extreme illumination.

\section{Method}
\label{sec:method}
\subsection{Overview}

As shown in Figure~\ref{fig:framework}, a pre-trained estimator first generates normal maps from RGB images. Concurrently, separate backbones extract multi-scale features from RGB frames and event streams. Fusion modules then integrate these geometric, appearance, and temporal cues. Finally, a detection head outputs bounding boxes from the fused representation. This design synergizes normal maps, events, and RGB textures for robust detection under low-light and motion-blur conditions.
\subsection{Monocular Normal Map Generation}
\label{sec:normal_gen}

To introduce robust geometric priors, we leverage DSINE~\cite{bae2024rethinking}, one of the state-of-the-art frameworks for estimating surface orientation from monocular RGB images. Unlike traditional approaches that derive normals from noisy depth gradients, DSINE directly regresses pixel-wise surface normals by learning to rotate a canonical representation based on image features. This inductive bias explicitly models the relationship between surface orientation and appearance, yielding superior generalization.

Formally, given an input RGB image $\mathbf{I} \in \mathbb{R}^{H \times W \times 3}$, we employ the estimator $\Phi_{\text{DSINE}}$ to predict the normal map $\mathbf{N} \in \mathbb{R}^{H \times W \times 3}$:
\begin{equation}
    \mathbf{N} = \Phi_{\text{DSINE}}(\mathbf{I}),
\end{equation}
where each pixel vector $\mathbf{n}_{uv} \in \mathbf{N}$ is unit-normalized, \ie, $\|\mathbf{n}_{uv}\|_2 = 1$.

\subsection{Event Representation}
To effectively process event camera data, we map the event data into structured representations. Specifically, the mapping process is formulated as:
\begin{equation}
E_i = K(x - x_n, y - y_n, t - t_n), \quad e_n \in \Delta T
\label{eq:event_kernel}
\end{equation}
where $K(x, y, t)$ is a kernel function, and $e_n$ represents the $n$-th event located within a given time window $\Delta T$. 
\subsection{Adaptive Dual-stream Fusion Module}
To robustly fuse complementary cues from RGB images and normal maps, we propose the \emph{Adaptive Dual-stream Fusion Module} (ADFM). First, a $1\times1$ convolution is applied to each modality to align feature dimensions. Given the RGB feature $\mathbf{F}^r \in \mathbb{R}^{H \times W \times C}$ and the normal map feature $\mathbf{F}^n \in \mathbb{R}^{H \times W \times C}$, we obtain:
\begin{equation}
\mathbf{F}^r_{\text{red}} = \operatorname{Conv}_{1\times1}(\mathbf{F}^r), \quad \mathbf{F}^n_{\text{red}} = \operatorname{Conv}_{1\times1}(\mathbf{F}^n),
\end{equation}
where $\mathbf{F}^r_{\text{red}}, \mathbf{F}^n_{\text{red}} \in \mathbb{R}^{H \times W \times C'}$ with $C' < C$.

Next, we reshape the feature maps by flattening spatial dimensions:
\begin{equation}
\tilde{\mathbf{F}}^r = T(\mathbf{F}^r_{\text{red}}) \in \mathbb{R}^{C' \times N}, \quad \tilde{\mathbf{F}}^n = T(\mathbf{F}^n_{\text{red}}) \in \mathbb{R}^{C' \times N},
\end{equation}
where $N = H \times W$ and $T(\cdot)$ denotes the flattening operation. To capture long-range dependencies, a cross-attention map is computed:
\begin{equation}
\mathbf{A} = \operatorname{Softmax}\Bigl(\tilde{\mathbf{F}}^{r\top}\tilde{\mathbf{F}}^n\Bigr) \in \mathbb{R}^{N \times N}.
\end{equation}
The normal features are aggregated as $\mathbf{F}^n_{\text{att}} = \tilde{\mathbf{F}}^n \cdot \mathbf{A}^\top$ and reshaped back to $\hat{\mathbf{F}}^n \in \mathbb{R}^{H \times W \times C'}$.

To refine the fused representation, a learnable scaling factor $\alpha$ controls the fusion strength. The final output is computed via a residual connection:
\begin{equation}
\mathbf{F}^r_{\text{fused}} = \mathbf{F}^r + \alpha \cdot \operatorname{Conv}_{1\times1}\bigl(\hat{\mathbf{F}}^n\bigr).
\end{equation}
This global cross-attention mechanism allows the network to selectively rely on geometric consistency when RGB appearance is ambiguous.

\subsection{Event-modality Aware Fusion Module}
We propose the \emph{Event-modality Aware Fusion Module} (EAFM) to integrate event features with the fused RGB--Normal features. Let $\mathbf{F}^{\mathrm{A}} \in \mathbb{R}^{C \times H \times W}$ denote the fused appearance/geometry features and $\mathbf{F}^{\mathrm{E}} \in \mathbb{R}^{C \times H \times W}$ the event features. EAFM first computes initial interactions:
\begin{equation}
\begin{aligned}
\mathbf{F}^{\mathrm{A+E}}_{0} &= \mathbf{F}^{\mathrm{A}} \odot \mathbf{F}^{\mathrm{E}} + \mathbf{F}^{\mathrm{A}},\\[1ex]
\mathbf{F}^{\mathrm{E+A}}_{0} &= \mathbf{F}^{\mathrm{A}} \odot \mathbf{F}^{\mathrm{E}} + \mathbf{F}^{\mathrm{E}},
\end{aligned}
\end{equation}
where $\odot$ denotes element-wise multiplication. These interactions are refined by sequentially applying convolutions and Group Normalization (GN). 

To generate adaptive spatial weights, we use global pooling followed by sigmoid activation:
\begin{equation}
W^{\mathrm{A+E}} = \sigma\Bigl(\operatorname{Conv}_{1\times1}\bigl(\operatorname{Pool}(\mathbf{F}^{\mathrm{A+E}}_{1})\bigr)\Bigr).
\end{equation}
The weighted features are obtained via $\mathbf{F}^{\mathrm{A+E}}_{\text{weighted}} = \mathbf{F}^{\mathrm{A+E}}_{1} \odot W^{\mathrm{A+E}}$. Finally, the streams are concatenated and adjusted:
\begin{equation}
\mathbf{F}^{\mathrm{EAFM}} = \operatorname{ChannelAdjust}\Bigl(\operatorname{Concat}\bigl(\mathbf{F}^{\mathrm{A+E}}_{\text{weighted}},\,\mathbf{F}^{\mathrm{E+A}}_{\text{weighted}}\bigr)\Bigr).
\end{equation}

\section{Experiments}

We evaluate NRE-Net on two challenging benchmarks: \textbf{PKU-DAVIS-SOD}~\cite{li2023sodformer} and \textbf{DSEC-Det-sub}~\cite{Gehrig21ral}. These datasets are selected for their specific inclusion of adverse illumination scenarios and high dynamic range events.

\subsection{Experimental Setup}
\paragraph{Datasets} \textbf{PKU-DAVIS-SOD}\cite{li2023sodformer} comprises 276k frames captured by a Davis346 camera ($346 \times 260$), featuring specific ``motion blur'' and ``low-light'' subsets. \textbf{DSEC-Det-sub}\cite{Gehrig21ral} contains 22k frames from a high-res Prophesee Gen3 camera ($640 \times 480$), offering a balanced mix of day and night scenes for robust evaluation.

\paragraph{Evaluation Metrics} 

We adopt the Average Precision at IoU=0.50 (denoted as \textbf{AP$_{\text{50}}$}) and the primary COCO metric averaged over IoU thresholds from 0.50 to 0.95 (denoted as \textbf{mAP}).

\subsection{Implementation Details} For our NRE-Net, we adopt YOLOX\cite{yolox2021} as the single-modality baseline model. The network is trained using Adam with an initial learning rate of \(1 \times 10^{-3}\). The input resolution is set to \(256 \times 320\) for the PKU-DAVIS-SOD dataset\cite{li2023sodformer} with a batch size of 16, while for the DSEC-Det-sub dataset\cite{Gehrig21ral}, the input resolution is \(480 \times 640\) with a batch size of 8.
\subsection{Quantitative Results}

Table~\ref{tab:comparison_results} compares NRE-Net against single-modality and fusion baselines.
\textbf{On DSEC-Det-sub}\cite{Gehrig21ral}, NRE-Net achieves \textbf{78.1\% AP$_{\text{50}}$} and \textbf{49.9\% mAP}, surpassing the strong fusion baseline SFNet~\cite{liu2024enhancing} by 2.7\%.
\textbf{On PKU-DAVIS-SOD}\cite{li2023sodformer}, our method achieves \textbf{57.5\% AP$_{\text{50}}$}, outperforming the transformer-based SODFormer (+7.1\%). This validates the efficacy of our hierarchical fusion strategy and the value of integrating monocular normal maps as an explicit geometric prior. 

\begin{table*}[hbtp!]
    \centering
    \caption{Quantitative detection performance on the DSEC-Det-sub\cite{Gehrig21ral} and PKU-DAVIS-SOD datasets\cite{li2023sodformer}.  Our proposed NRE-Net consistently outperforms competing approaches in terms of AP$_{50}$(\%) and mAP(\%).}
    \label{tab:comparison_results}
    \begin{tabular}{c c c c c c c c}
    \toprule
    \multirow{2}{*}{\textbf{Model Type}} & \multirow{2}{*}{\textbf{Method}} & \multirow{2}{*}{\textbf{Pub. \& Year}} & \multirow{2}{*}{\textbf{Input}} 
    & \multicolumn{2}{c}{\textbf{DSEC-Det-sub}} 
    & \multicolumn{2}{c}{\textbf{PKU-DAVIS-SOD}} \\ 
    \cmidrule(lr){5-6} \cmidrule(lr){7-8}
    & & & & \textbf{AP$_{50}$} & \textbf{mAP} 
    & \textbf{AP$_{50}$} & \textbf{mAP} \\
    \midrule
    
    \multirow{5}{*}{Event-ONLY} 
      & YOLOX\cite{yolox2021}     & arXiv'2021  & Event Frame   & 50.7 & 23.3  & 43.7 & 20.5 \\
      & RVT\cite{Gehrig_2023_CVPR}       & CVPR'2023  & Voxel         & 50.9 & 26.0  & 44.1 & 20.3 \\
      & DMANet\cite{wang2023dual}    & AAAI'2023  & EventPillars  & 36.5 & 19.2  & 30.6   & 12.1   \\
      & SAST\cite{peng2024scene}      & CVPR'2024  & Voxel         & 52.5 & 27.8  & 46.2 & 22.1 \\
      & SMamba\cite{yang2025smamba}    & AAAI'2025 & Voxel         & 54.2 & 29.1  & 47.9 & 23.5 \\[0.5em]
      
    \multirow{4}{*}{RGB-ONLY}
      & RetinaNet\cite{lin2017focal} & CVPR'2017    & RGB           & 61.7 & 36.6  & 39.2 & 13.9 \\
      & YOLOv5\cite{jocher2020ultralytics}    & arXiv'2020         & RGB           & 69.8 & 41.6  & 50.2 & 22.6 \\
      & YOLOX\cite{yolox2021}     & arXiv'2021  & RGB           & 70.2 & 43.5  & 51.4 & 24.3\\
      & YOLOv7\cite{wang2023yolov7}    & CVPR'2023         & RGB           & 72.6 & 45.7  & 52.7 & 26.4 \\[0.5em]
      
    \multirow{6}{*}{Multi-modal}
      & FPN-fusion\cite{tomy2022fusing} & ICRA'2022  & RGB, Voxel     & 63.7 & 36.9  & 40.3 & 15.2 \\
      & RENet\cite{zhou2023rgb}     & ICRA'2023  & RGB, E-TMA     & 64.9 & 38.6  & 42.5   & 18.7   \\
      & SODFormer\cite{li2023sodformer} & TPAMI'2023  & RGB, Event Frame & --   & --   & 50.4 & 20.7 \\
      & EOLO\cite{cao2024chasing}      & ICRA'2024  & RGB, Event Frame & 68.1 & 41.6  & 44.8 & 19.4 \\
      & SFNet\cite{liu2024enhancing}     & TITS'2024  & RGB, Event Frame & 75.4 & 46.7  & 54.3 & 25.7 \\
      & \textbf{NRE-Net (Ours)} 
                  & ICME'2026   & RGB, Event Frame & \textbf{78.1} & \textbf{49.9} & \textbf{57.5} & \textbf{29.0} \\
    \bottomrule
    \end{tabular}
\end{table*}

\subsection{Ablation Studies}

\subsubsection{Impact of Modality Composition}
To quantify the explicit contribution of geometric priors, we perform a component-wise ablation on DSEC-Det-sub~\cite{Gehrig21ral} and PKU-DAVIS-SOD~\cite{li2023sodformer}, as detailed in Table~\ref{tab:ablation_modal_pku}.
While single modalities struggle in adverse conditions (Event-ONLY: 50.7\%, RGB-ONLY: 70.2\%), the dual-modal RGB+Event configuration establishes a robust baseline of 75.1\% by combining texture and dynamics.
Crucially, integrating the Normal Map branch further elevates performance to \textbf{78.1\%}. This specific \textbf{3.0\% gain} over the strong dual-modal baseline is significant in high-precision regimes.

\begin{table}[hbtp!]
    \centering
    \caption{Impact of Modality Composition. Component-wise ablation quantifying the contribution of each modality.}
    \label{tab:ablation_modal_pku}
    \small  
    \begin{tabular}{l cc cc}
        \toprule
        \multirow{2}{*}{\textbf{Input Modalities}} & 
        \multicolumn{2}{c}{\textbf{DSEC-Det-sub}} & \multicolumn{2}{c}{\textbf{PKU-DAVIS}} \\
        \cmidrule(lr){2-3} \cmidrule(lr){4-5}
        & \textbf{AP$_{\text{50}}$} & \textbf{mAP} & \textbf{AP$_{\text{50}}$} & \textbf{mAP} \\
        \midrule
        \multicolumn{5}{l}{\textit{I. Single Modality}} \\
        Event-ONLY          & 50.7 & 23.3 & 43.7 & 20.5 \\
        RGB-ONLY            & 70.2 & 43.5 & 51.4 & 24.3 \\
        \midrule
        \multicolumn{5}{l}{\textit{II. Dual Modalities}} \\
        Event + Normal      & 61.3 & 35.9 & 51.1 & 25.3 \\
        RGB + Normal        & 72.0 & 46.9 & 52.5 & 25.6 \\
        RGB + Event \textit{(Baseline)} & 75.1 & 46.2 & 55.6 & 26.4 \\
        \midrule
        \multicolumn{5}{l}{\textit{III. Ours}} \\
        \textbf{RGB + Event + Normal} & \textbf{78.1} & \textbf{49.9} & \textbf{57.5} & \textbf{29.0} \\
        \bottomrule
    \end{tabular}
\end{table}

\subsubsection{Robustness to RGB Degradation}
To assess the dependency of our method on the quality of the input RGB image, from which normal maps are derived, we conducted a targeted analysis on the challenging subsets of PKU-DAVIS-SOD~\cite{li2023sodformer}. 
As detailed in Table~\ref{tab:pku-davis-sod-combined}, in low-light scenarios where RGB texture is severely compromised, NRE-Net still achieves \textbf{47.0\% AP$_{\text{50}}$}, outperforming the RGB-ONLY baseline (41.9\%) and the RGB+Event baseline (45.2\%). 

We further provide qualitative visualizations in the Supplemental Material to substantiate this resilience. Since normal maps primarily encode low-frequency structural geometry (e.g., planar surfaces) rather than high-frequency texture details, they remain robust in most typical scenarios, such as low-light or overexposure. Consequently, even degraded RGB inputs can yield structurally meaningful geometric cues that effectively assist event-based object detection.

\begin{table}[hbtp!]
    \centering
    \caption{Detection Performance on the Motion Blur and Low Light subsets of PKU-DAVIS-SOD. }
    \label{tab:pku-davis-sod-combined}
    \resizebox{\columnwidth}{!}{
    \begin{tabular}{l l cc}
        \toprule
        \multirow{2}{*}{\textbf{Condition}} & \multirow{2}{*}{\textbf{Input Modalities}} & \multicolumn{2}{c}{\textbf{Performance}} \\
        \cmidrule(lr){3-4}
        & & \textbf{AP$_{\text{50}}$} & \textbf{mAP} \\
        \midrule
        \multirow{3}{*}{\textit{Motion Blur}} 
        & RGB-ONLY & 48.2 & 22.7 \\ 
        & RGB+Event & 48.7 & 24.7 \\ 
        & \textbf{RGB+Event+Normal} & \textbf{49.1} & \textbf{25.0} \\ 
        \midrule
        \multirow{3}{*}{\textit{Low Light}} 
        & RGB-ONLY & 41.9 & 18.0 \\ 
        & RGB+Event & 45.2 & 20.1 \\ 
        & \textbf{RGB+Event+Normal} & \textbf{47.0} & \textbf{21.2} \\ 
        \bottomrule
    \end{tabular}
    }
\end{table}

\subsubsection{Ablation and Comparison of Fusion Mechanisms}
To verify the necessity of our hierarchical attention design, we benchmark NRE-Net against standard fusion paradigms. As presented in Table~\ref{tab:fusion_ablation}, generic operations yield limited performance gains (plateauing at $\sim$75.5\% on DSEC-Det-sub).

In contrast, our specialized modules confer improvements. As shown in the Gain column of Table~\ref{tab:fusion_ablation} (computed relative to the Naive Sum baseline on DSEC-Det-sub), introducing \textbf{ADFM} yields a \textbf{1.6\%} AP$_{\text{50}}$ increase, while \textbf{EAFM} contributes \textbf{2.3\%}. Ultimately, the full \textbf{NRE-Net} achieves \textbf{78.1\%} AP$_{\text{50}}$, establishing a clear \textbf{3.2\%} margin over the baseline. This validates that adaptive, geometry-aware fusion is essential for resolving inter-modal conflicts in complex illumination scenarios.
\begin{table}[hbtp!]
    \centering
    \caption{Analysis of Fusion Mechanisms. We benchmark our proposed modules against generic fusion strategies. \textbf{Naive Sum} (Element-wise Addition) serves as the reference baseline. The \textbf{Gain} column indicates the absolute improvement in AP$_{\text{50}}$ over this baseline on the DSEC-Det-sub.}

    \label{tab:fusion_ablation}
    \small
    
    \begin{tabular}{l cc cc c}
        \toprule
        \multirow{2}{*}{\textbf{Method}} & 
        \multicolumn{2}{c}{\textbf{DSEC-Det-sub}} & 
        \multicolumn{2}{c}{\textbf{PKU-DAVIS}} & 
        \multirow{2}{*}{\textbf{Gain}} \\
        \cmidrule(lr){2-3} \cmidrule(lr){4-5}
        & \textbf{AP$_{\text{50}}$} & \textbf{mAP} & 
        \textbf{AP$_{\text{50}}$} & \textbf{mAP} & 
        \\
        \midrule
        \multicolumn{6}{l}{\textit{\textbf{I. Baselines}}} \\
        Naive Sum & 74.9 & 46.4 & 55.4 & 27.0 & \textit{Ref.} \\
        Max Pooling & 74.4 & 46.6 & 56.4 & 27.9 & -0.5 \\
        MLP-Fusion & 75.5 & 46.7 & 55.3 & 27.2 & +0.6 \\
        \midrule
        \multicolumn{6}{l}{\textit{\textbf{II. Proposed NRE-Net (Ablation)}}} \\
        w/ ADFM & 76.5 & 47.7 & 57.4 & 28.6 & +1.6 \\
        w/ EAFM & 77.2 & 48.5 & 56.3 & 28.5 & +2.3 \\
        \textbf{Full Model} & \textbf{78.1} & \textbf{49.9} & \textbf{57.5} & \textbf{29.0} & \textbf{+3.2} \\
        \bottomrule
    \end{tabular}
\end{table}

\subsubsection{Fusion-Module Placement}
Finally, we investigate the optimal stages for injecting our fusion modules. We integrate the proposed modules at the F3, F4, and F5 feature outputs of the CSPDarkNet backbone (corresponding to \textit{layer2}, \textit{layer3}, and \textit{layer4}). 
As shown in Table~\ref{tab:fusion_placement}, applying fusion solely at the deepest semantic level (\textit{layer4}) yields 76.7\% AP$_{\text{50}}$. Extending fusion to include \textit{layer3} further improves performance to 77.9\%. 
The optimal performance (78.1\%) is achieved when fusion is applied across all three scales (\textit{layer2, 3, 4}). This demonstrates that while high-level semantic features are crucial, incorporating multi-scale representations is essential for maximizing detection robustness in complex scenarios.

\begin{table}[hbtp!]
    \centering
    \caption{Ablation of Fusion Placement. Performance comparison across different backbone depths. \textit{Layer 2-4} correspond to the F3-F5 outputs of the feature extractor. }
    
    \label{tab:fusion_placement}
    \resizebox{\columnwidth}{!}{
    \begin{tabular}{ccc|cc|cc}
        \toprule
        \multicolumn{3}{c|}{\textbf{Fusion Stages}} & 
        \multicolumn{2}{c|}{\textbf{DSEC-Det-sub}} & 
        \multicolumn{2}{c}{\textbf{PKU-DAVIS}} \\
        \textbf{\textit{layer2}} & \textbf{\textit{layer3}} & \textbf{\textit{layer4}} & 
        \textbf{AP$_{\text{50}}$} & \textbf{mAP} & 
        \textbf{AP$_{\text{50}}$} & \textbf{mAP} \\
        \midrule
        - & - & - & 74.9 & 46.4 & 55.4 & 27.0 \\
        - & - & \checkmark & 76.7 & 48.6 & 56.3 & 28.4 \\
        - & \checkmark & \checkmark & 77.9 & 49.5 & \textbf{57.8} & 29.0 \\
        \checkmark & \checkmark & \checkmark & \textbf{78.1} & \textbf{49.9} & 57.5 & \textbf{29.0} \\
        \bottomrule
    \end{tabular}
    }
\end{table}

\subsection{Qualitative Analysis}

\begin{figure}[htbp!]
    \centering
    \includegraphics[trim=0.65cm 10.2cm 4.7cm 2.5cm, clip, width=0.49\textwidth, height=8.8cm]{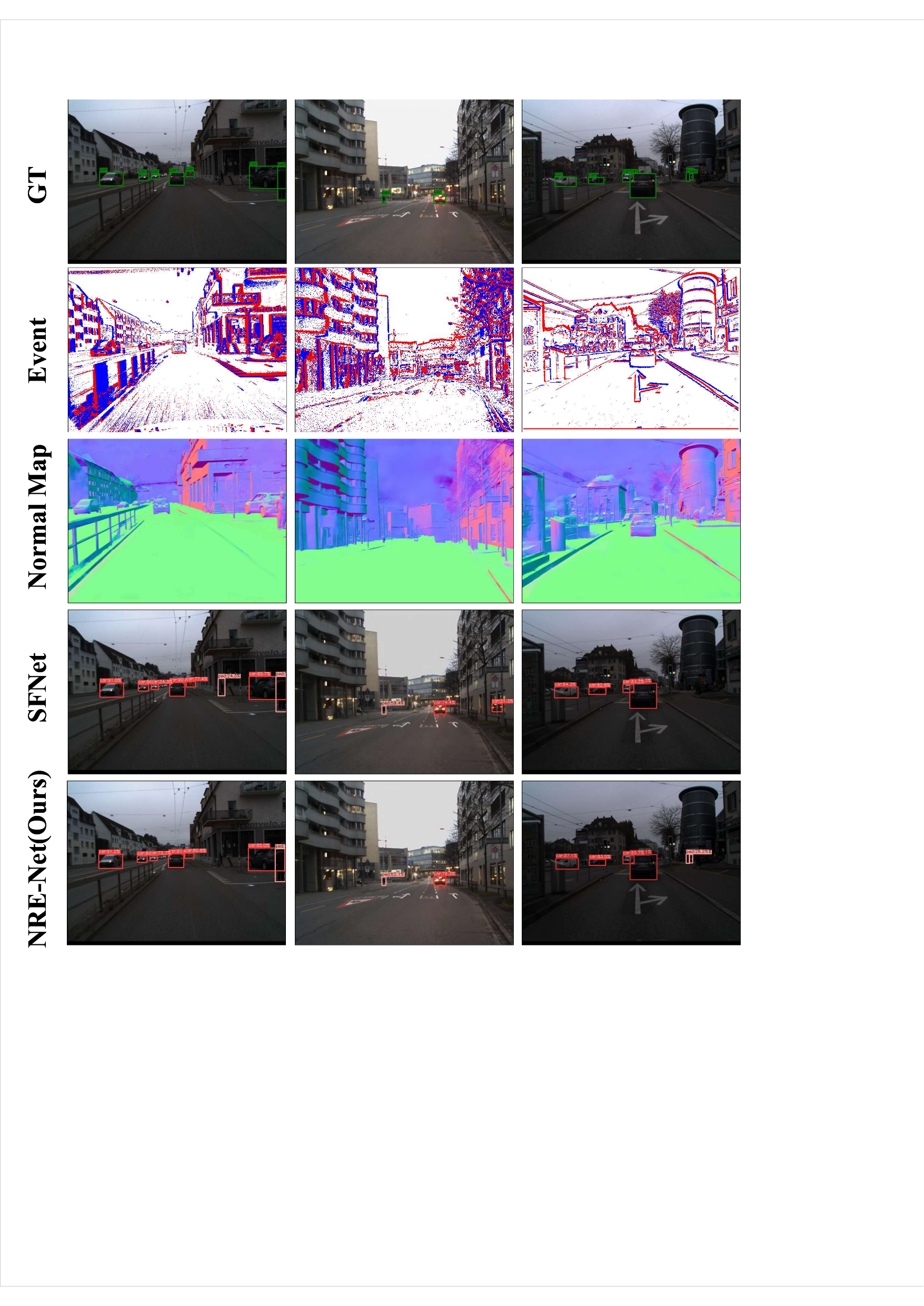}
 \caption{Visual comparison: GT, Events, Normals, SFNet, Ours.}

    \label{fig:comparasion}
\end{figure}

Fig.~\ref{fig:comparasion} visualizes detection performance under various illumination scenarios. NRE-Net produces more precise and reliable detection results while mitigating the critical limitations of methods that rely solely on appearance and motion cues.

\section{Conclusion}
\label{sec:conclusion}
In this work, we present NRE-Net, a trimodal framework leveraging monocular surface normal maps as explicit geometric priors to resolve perceptual ambiguities in event-based detection. By hierarchically fusing appearance, dynamics, and geometry, our approach effectively enhances the robustness of event-based detection under complex illumination. Extensive evaluations on DSEC-Det-sub and PKU-DAVIS-SOD demonstrate the efficacy of incorporating geometric constraints for robust perception in adverse environments.
\section*{Acknowledgment}
This work was supported by the Beijing Natural Science Foundation (4262060), the National Key R\&D Program of China (2021ZD0109802), and the High-performance Computing Platform of BUPT.

\bibliographystyle{IEEEbib}
\bibliography{icme2025references}

\vspace{12pt}

\end{document}